\documentclass[letterpaper, 10 pt, conference]{ieeeconf}

\IEEEoverridecommandlockouts
\usepackage{booktabs}
\usepackage{graphicx}
\usepackage{amsmath}
\usepackage{amssymb}
\usepackage{amsfonts}
\usepackage{algorithmic}
\usepackage{array}
\usepackage{enumerate}
\usepackage{textcomp}
\usepackage{stfloats}
\usepackage{url}
\usepackage{verbatim}
\usepackage{caption}
\usepackage{subcaption}
\usepackage[ruled,vlined,linesnumbered]{algorithm2e}
\usepackage{cite}
\usepackage{bm}
\usepackage{tikz}
\usepackage{color}
\usepackage{mathtools}
\usepackage{hyperref}
\hypersetup{
    colorlinks=true,
    linkcolor=blue,
    filecolor=blue,
    citecolor=blue,
    urlcolor=blue,
}
\usepackage{rotating}
\usepackage{blkarray}
\usepackage{physics}
\usetikzlibrary{arrows}

\usepackage{amsthm}
\usepackage{comment}
\usepackage{soul}
\usepackage{balance}

\title{\LARGE \bf
From Sketch Prior to Trajectories: A Mission-Oriented Coordinated Navigation Framework for Indoor UAV Swarm
}

\author{Xinhang~Xu$^{*}$, Ruiyang~Liu$^{*}$, Haotian~Jin$^{*}$,  Yi~Wang, Hongming~Shen,\\ Jianping Li, \IEEEmembership{Senior Member,~IEEE}, Lihua~Xie$^{1}$, \IEEEmembership{Fellow,~IEEE}\thanks{$^{*}$Equal Contribution.}
\thanks{Xinhang Xu, Haotian Jin, Yi Wang, Hongming Shen, and Jianping Li are with the School of Electrical and Electronic Engineering, Nanyang Technological University, 50 Nanyang Avenue, Singapore. Lihua Xie is with NTU–VinUni Joint Research Laboratory for Robotics and Embodied AI, School of Electrical and Electronic Engineering, Nanyang Technological University, Singapore 639798 and VinUniversity, Hanoi 100000, Vietnam}
\thanks{$^{1}$Corresponding author. Email: {\tt elxie@ntu.edu.sg}. }
}

\begin{document}

\maketitle

\begin{abstract}
Multi-UAV indoor operations, such as inspection, security patrol, and logistics delivery, are often mission-oriented rather than exploration-oriented. In these tasks, UAVs are typically required to visit task-relevant regions in a prescribed sequence, and such region-level mission information can be obtained from pre-deployment sketch-map priors, including floor plans, CAD layouts, and evacuation diagrams. Although the tasks are executed in three-dimensional space, UAVs usually operate within a limited altitude range on each floor, such that mission-level transitions are largely governed by planar connectivity. Motivated by these characteristics, this paper proposes a mission-oriented coordinated navigation framework using sketch-map priors. Onboard observations are used to align the prior and update a mission-oriented traversability representation, while a layered 2D--3D planner generates mission-oriented guide paths and refines them into dynamically feasible and collision-free trajectories. Simulation and real-world experiments in structured multi-room environments demonstrate the effectiveness of the proposed framework under both communication-available and communication-loss conditions. Multi-floor simulations further demonstrate its scalability to layered indoor structures.
\end{abstract}

\section{Introduction}


Multi-UAV systems have shown significant potential in indoor inspection, emergency response, security patrol, and logistics operations. Recent advances in aerial autonomy have substantially improved the maneuverability of individual platforms, enabling precise and aggressive flight in confined spaces~\cite{tianyue2026precise}. Beyond free flight, recent systems enable whole-body aerial manipulation~\cite{jin2026quadhand}, high-force push-and-pull interactions~\cite{wang2026ductam}, and tethered transport through autonomous knotting~\cite{jin2025tethered}. Through cooperative execution, heterogeneous sensing, and complementary payloads, multiple UAVs can address mission objectives that are difficult for a single platform. However, practical indoor environments are typically GNSS-denied and consist of interconnected rooms, corridors, stairwells, and functional spaces, making efficient and coordinated mission execution challenging.


Existing UAV navigation research is largely environment-centric. Exploration methods~\cite{geng2025epic,li2026aeos,zacharia2026omniplanner,li2026fu,tang2023bubble} focus on recovering unknown environmental structures through frontier-based planning, information-gain maximization, and cooperative mapping, while coverage methods~\cite{xu2024cost,cao2023distributed,wu2026layer,jin2024gs,cao2020online} aim to visit reachable regions for applications such as inspection and environmental monitoring. These methods primarily emphasize environmental completeness. In many indoor operations, however, the mission is instead specified by task-relevant regions and their execution order~\cite{hong2026post,cao2025cooperative,pan2026action}, such as designated patrol areas, emergency response locations, or delivery destinations. Directly applying environment-centric methods may therefore introduce task-irrelevant sensing and motion.


Many structured indoor environments provide useful pre-deployment priors~\cite{wang2026palm,dharmadhikari2025semantics,baek2025pipe}, including architectural layouts, floor plans, CAD models, evacuation maps, and sketch maps. Such priors can describe room geometry, region connectivity, entrances, exits, and task-relevant regions, making them well suited to region-level mission specifications. However, they are not directly executable navigation maps because map-to-world misalignment, local environmental changes, static and dynamic obstacles, and inter-UAV interactions still need to be handled online. The resulting challenge is to convert the sketch-map prior into a navigation-compatible representation that can be aligned and updated during execution.


A further observation is that indoor building operations often exhibit a layered, 2.5D structure~\cite{border2024osprey}. Although UAVs operate in three-dimensional space, they typically maintain a relatively stable altitude within each floor, such that mission-level transitions are largely governed by the planar connectivity of rooms and corridors. Vertical motion mainly occurs at designated inter-floor transitions, such as staircases, elevators, or atriums. This structure allows mission-level reasoning, including target switching, region-state coordination, and multi-UAV conflict resolution, to be performed in a two-dimensional abstraction, while local three-dimensional representations are retained for collision checking and trajectory generation. This observation motivates a layered architecture combining 2D guidance with 3D trajectory optimization.


Based on these observations, this paper proposes a mission-oriented coordinated navigation framework for multi-UAV indoor operations using sketch-map priors. Rather than relying on a shared dense global metric map, the framework treats the sketch map as a lightweight structural prior and aligns it with online observations to construct a mission-oriented traversability representation integrating static structure, dynamic traversability, and region states. A layered 2D--3D navigation architecture then generates mission-oriented guide paths and refines them into dynamically feasible and collision-free trajectories, linking region-level mission guidance with local motion planning. The main contributions are summarized as follows:

\begin{figure*}[t]
    \centering
    \includegraphics[width=0.9\textwidth]{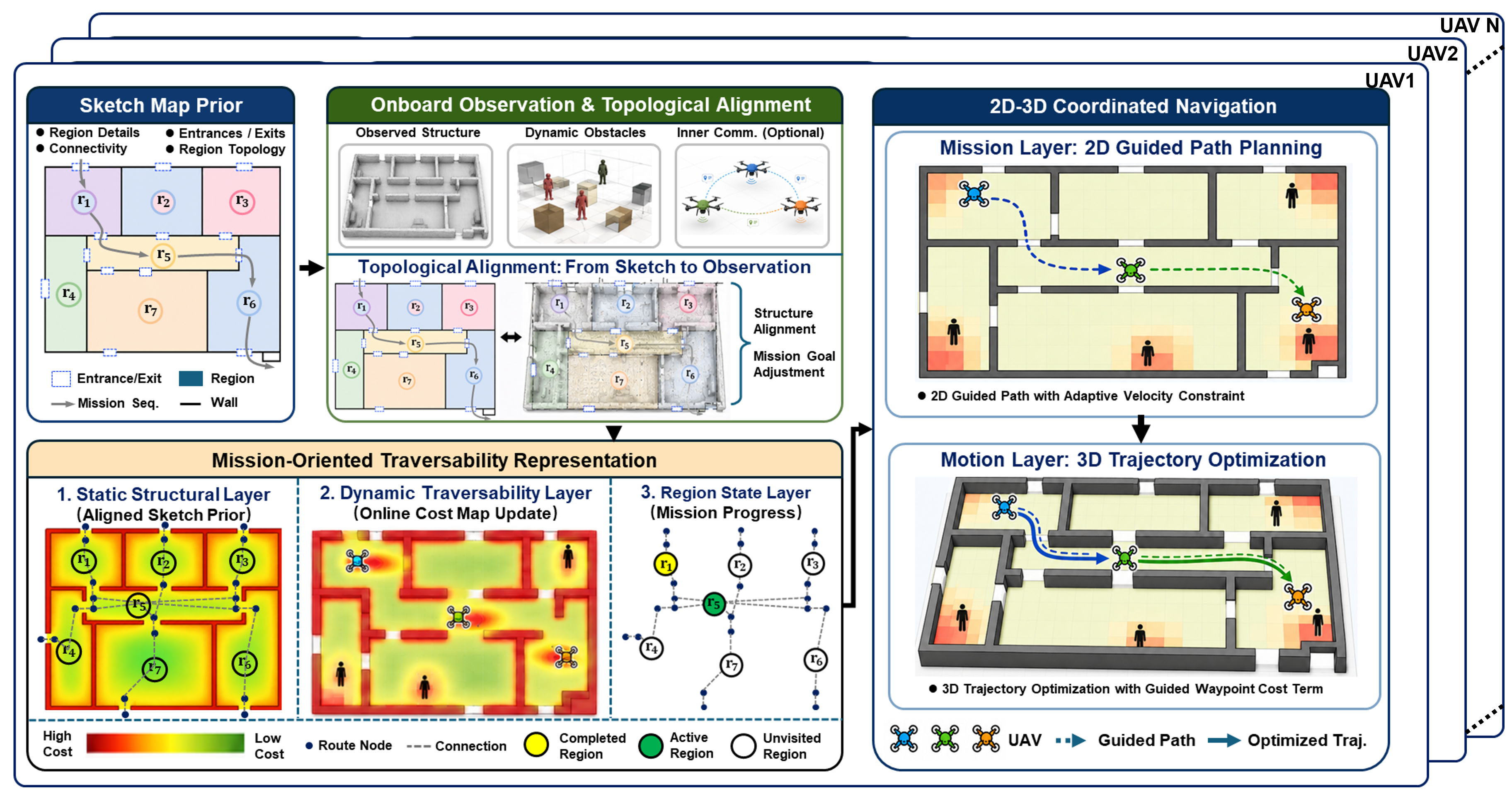}
    \caption{Overview of the proposed mission-oriented coordinated navigation framework for multi-UAV building operations. A sketch-map prior provides coarse region topology, connectivity, entrances/exits, and mission sequence information. During operation, onboard observations are used for topological alignment and dynamic traversability updates. The resulting mission-oriented traversability representation consists of a static structural layer, a dynamic traversability layer, and a region-state layer. A 2D guided path planning layer generates mission-oriented guide paths, while a guide-driven 3D trajectory optimization layer refines the guidance into dynamically feasible and collision-free UAV trajectories using local 3D occupancy maps.}
    \label{fig:system_overview}
\end{figure*} 

\begin{itemize}
\item We propose a mission-oriented traversability representation that integrates region-level topological alignment, static structural constraints, dynamic traversability, and region-state information, providing a compact navigation interface without requiring dense global maps.
\item We develop a layered 2D--3D coordinated navigation framework that leverages the 2.5D structure of building operations. The 2D layer provides mission-level guidance over planar region connectivity and dynamic traversability, while the 3D layer refines the guidance into dynamically feasible and collision-free trajectories.
\item We validate the framework through simulation and real-world multi-UAV experiments in structured multi-room environments under both communication-available and communication-loss conditions. Additional multi-floor simulations show its scalability to layered structures.
\end{itemize}

\section{Problem Statement and System Overview}

\subsection{Problem Statement}

Consider a team of $N$ UAVs operating in a GNSS-denied indoor environment. A pre-deployment sketch map is represented as a topological graph $\mathcal{M}=(\mathcal{R},\mathcal{E})$, where $\mathcal{R}=\{r_1,r_2,\ldots,r_K\}$ denotes $K$ mission regions, such as rooms, corridors, stairwells, and halls, and $\mathcal{E}$ denotes traversable connections between neighboring regions. The sketch map serves as a structural prior that provides region geometry, topology, and connectivity, while map-to-world alignment and local changes are handled online. The mission is specified by an ordered sequence of $M$ target regions, $\mathcal{G}=(g_1,g_2,\ldots,g_M)$, with $g_j\in\mathcal{R}$ defining the mission progression. Each region maintains a mission state
\begin{equation}
s(r)\in\{\texttt{unvisited},\texttt{active},\texttt{completed}\},
\label{eq:region_state}
\end{equation}
to represent mission progress and support team coordination. Let $\mathcal{F}(t)$ denote the dynamically traversable free space at time $t$, and $d_{\mathrm{safe}}$ the minimum inter-UAV safety distance.

The objective is to generate trajectories
\(\mathcal{T}=\{\tau_1,\tau_2,\ldots,\tau_N\}\)
that complete the specified mission sequence while satisfying environmental, inter-UAV safety, and region-order constraints. At the mission level, this objective is expressed by the following formulation:
\begin{equation}
\begin{aligned}
\min_{\mathcal{T},T_f}
\quad &
T_f + \lambda \sum_{i=1}^{N}\int_{0}^{T_f}
\|\dot{\mathbf{p}}_i(t)\| \, dt \\
\text{s.t.}
\quad &
\mathbf{p}_i(t)\in\mathcal{F}(t),
\qquad \forall i, \\
&
\|\mathbf{p}_i(t)-\mathbf{p}_j(t)\|
\ge d_{\mathrm{safe}},
\qquad \forall i\neq j, \\
&
s(g_k)=\texttt{completed},
\qquad \forall g_k\in\mathcal{G}, \\
&
t(g_n)<t(g_{n+1}),
\qquad n=1,\ldots,M-1,
\end{aligned}
\label{eq:problem}
\end{equation}
where \(T_f\) is the mission completion time, \(\mathbf{p}_i(t)\) is the position of UAV \(i\), and \(t(g_n)\) denotes the completion time of target region \(g_n\). The objective balances mission completion time and total trajectory length through the weight \(\lambda\ge0\).

This formulation serves as a mission-level reference objective rather than a globally solved multi-agent optimal control problem. The proposed framework instead generates receding-horizon solutions through mission-oriented 2D guidance and local 3D trajectory optimization, where the former provides region-level routes and the latter refines them into dynamically feasible and collision-free trajectories.

\subsection{System Overview}

As shown in Fig.~\ref{fig:system_overview}, the proposed framework consists of three components: sketch-map alignment, mission-oriented traversability representation, and layered 2D--3D navigation. During operation, onboard observations are aligned with the sketch prior to identify the current mission region and valid neighboring transitions.
The aligned prior and online observations are integrated into a mission-oriented traversability representation comprising a static structural layer, a dynamic traversability layer, and a region-state layer. These layers encode structural constraints, time-varying environmental risks, and mission progress, respectively.

Based on this representation, the 2D layer generates a mission-aware guide toward the active target region, while the 3D layer refines a local guide segment into a dynamically feasible and collision-free trajectory using onboard occupancy information. Each UAV maintains its own local 3D occupancy map, while only lightweight region states and optional neighboring-UAV information are exchanged.

\section{Mission-Oriented Traversability Representation}

The mission-oriented traversability representation integrates the aligned sketch prior, online environmental observations, and region-level mission states into a unified navigation interface. It consists of a static structural layer, a dynamic traversability layer, and a region-state layer, which respectively encode persistent structural constraints, time-varying navigation risks, and mission progress.

\subsection{Static Structural Layer and Topological Alignment}

The sketch-map prior $\mathcal{M}=(\mathcal{R},\mathcal{E})$ provides region geometry and connectivity. During operation, local structural observations are aligned with the sketch prior using an ICP-based procedure to identify the current mission region and retrieve valid neighboring transitions.

The aligned prior is converted into a static structural layer $C_{\mathrm{struct}}$, which encodes walls, traversable doors, corridors, and inter-region connections. Areas unsupported by the aligned prior are assigned non-traversable costs to prevent the 2D guide from using invalid shortcuts.

\subsection{Dynamic Traversability Layer}

The static structural layer captures persistent building constraints, while online observations are used to model time-varying risks from moving obstacles and neighboring UAVs. The overall traversability cost is defined as
\begin{equation}
C_{\mathrm{trav}}(\mathbf{x},t)
=
C_{\mathrm{struct}}(\mathbf{x})
+
C_{\mathrm{dyn}}(\mathbf{x},t),
\label{eq:trav_cost}
\end{equation}
where $C_{\mathrm{dyn}}$ denotes the online dynamic cost.

Detected obstacles are inserted into the grid map using onboard point clouds and, when available, neighboring-UAV information. The obstacle-occupancy cost \(C_{\text{obs}}(x,t)\) is assigned \(C_{\text{lethal}}\) at occupied cells, with an exponentially decaying inflation field applied around them:
\begin{equation}
C_{\mathrm{inf}}(\mathbf{x},t)
=
\beta \exp\left[-\alpha\left(d(\mathbf{x},t)-r_{\mathrm{ins}}\right)\right],
\quad d(\mathbf{x},t) \leq r_{\mathrm{inf}},
\label{eq:inflation_cost}
\end{equation}
where $d(\mathbf{x},t)$ is the distance to the nearest occupied cell, $\beta$ and $\alpha$ control the inflation magnitude and decay rate, and $r_{\mathrm{ins}}$ and $r_{\mathrm{inf}}$ denote the inscribed and inflation radius.
The occupancy and inflation terms form the base dynamic cost:
\begin{equation}
C_{\mathrm{base}}(\mathbf{x},t)
=
\min\left(
C_{\mathrm{lethal}},
C_{\mathrm{obs}}(\mathbf{x},t)
+
C_{\mathrm{inf}}(\mathbf{x},t)
\right).
\label{eq:base_cost}
\end{equation}

To capture evolving local risk, the temporal variation of the base cost is evaluated as
\begin{equation}
\Delta C_{\mathrm{base}}(\mathbf{x},t)
=
C_{\mathrm{base}}(\mathbf{x},t)
-
C_{\mathrm{base}}(\mathbf{x},t-1).
\label{eq:cost_variation}
\end{equation}
Within a sliding window of length $W$, the trend score is computed as
\begin{equation}
S(\mathbf{x},t)
=
\sum_{\tau=t-W+1}^{t}
k_d \Delta C_{\mathrm{base}}(\mathbf{x},\tau),
\label{eq:trend_score}
\end{equation}
and the corresponding trend cost is
\begin{equation}
C_{\mathrm{trend}}(\mathbf{x},t)
=
\max\left(0,k_t\bar{S}(\mathbf{x},t)\right),
\label{eq:trend_cost}
\end{equation}
where $k_d$ and $k_t$ are weighting coefficients and $\bar{S}$ denotes the average valid trend score within the window.

The final dynamic cost is
\begin{equation}
C_{\mathrm{dyn}}(\mathbf{x},t)
=
\min\left(
C_{\mathrm{lethal}},
C_{\mathrm{base}}(\mathbf{x},t)
+
C_{\mathrm{trend}}(\mathbf{x},t)
\right).
\label{eq:dyn_cost}
\end{equation}

This formulation captures both obstacle proximity and increasing local risk, and is used by the 2D guide planner as well as the adaptive velocity design introduced in Sec.~IV-B.

\subsection{Region-State Layer}

The region-state layer maintains mission progress using the region states defined in \eqref{eq:region_state}. According to the predefined mission sequence, completion of the active region triggers activation of the next target region, enabling region-level mission switching without requiring synchronization of dense maps or full trajectories.

When communication is available, UAVs exchange region states and optional coarse neighboring-UAV information, which can be incorporated into the dynamic traversability layer for conflict avoidance. Under communication loss, neighboring UAVs are handled through onboard perception as dynamic obstacles. If the communication-loss duration exceeds a predefined threshold, each UAV continues mission execution based on its local region state and observations.

Together, the three layers provide a compact mission-oriented representation for the subsequent 2D--3D coordinated navigation framework.

\section{2D--3D Coordinated Navigation Framework}

The coordinated navigation framework converts the mission-oriented traversability representation into executable UAV trajectories. The 2D layer generates a mission-aware guide on the traversability costmap, while the 3D layer refines the guide into a dynamically feasible and collision-free local trajectory.
\begin{figure*}[t]
    \centering
    \includegraphics[width=0.95\textwidth]{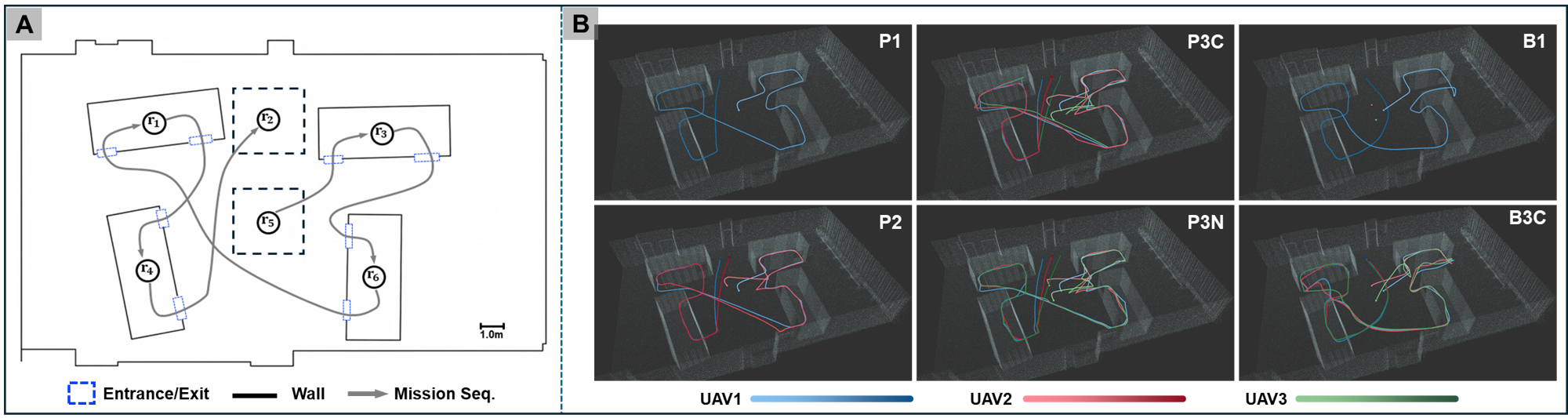}
    \caption{Comparative experimental setup and trajectory results.
    (A) Structured multi-room environment with six regions $r_1$--$r_6$ and the predefined mission sequence.
    (B) Trajectories of the fastest run among ten trials for the six comparative settings. Different colors represent different UAVs, while color intensity indicates temporal progression.}
    \label{fig:comparative_experiments}
\end{figure*}
\subsection{2D Guided Path Planning}

The 2D guide is generated on a grid-based traversability costmap. For each cell $v_i$, the planning cost is defined as
\begin{equation}
C_{\mathrm{plan}}(v_i,t)
=
w_s C_{\mathrm{struct}}(v_i)
+
w_d C_{\mathrm{dyn}}(v_i,t)
+
c_{\mathrm{n}},
\label{eq:planning_cost_raw}
\end{equation}
where $w_s$ and $w_d$ are the weights of the static structural and dynamic traversability costs, respectively, and $c_{\mathrm{n}}$ is the neutral traversal cost. The cost is saturated by the lethal threshold:
\begin{equation}
\bar{C}(v_i,t)
=
\min\left(
C_{\mathrm{plan}}(v_i,t),
C_{\mathrm{lethal}}
\right).
\label{eq:planning_cost}
\end{equation}

Given the current UAV cell $v_s$ and the active target region $g_j$, the goal cell $v_g$ is selected from its valid entrance, exit, or task location. A weighted A* search is then performed using
\begin{equation}
f(v)
=
c_{\mathrm{acc}}(v)
+
\epsilon h(v),
\label{eq:wastar}
\end{equation}
where $c_{\mathrm{acc}}(v)$ is the accumulated traversal cost, $h(v)$ is the heuristic distance to $v_g$, and $\epsilon\geq1$ controls the search greediness. Cells with lethal structural or dynamic costs are excluded from the search.

After planning, redundant waypoints are removed and the path is smoothed while preserving valid structural transitions. The resulting guide path,
\begin{equation}
\mathcal{P}_{\mathrm{2D}}
=
\{\mathbf{q}_0,\mathbf{q}_1,\ldots,\mathbf{q}_{N_q}\},
\qquad
\mathbf{q}_m=(x_m,y_m),
\label{eq:2d_guide_path}
\end{equation}
serves as a mission-level reference for local 3D trajectory optimization rather than being directly executed.

\subsection{Adaptive Velocity Constraint Mechanism}

The dynamic traversability layer is further used to adapt the local motion constraints according to forward risk. A set of points $S_p$ is sampled along the forward segment of the current 2D guide, and the maximum dynamic cost is defined as
\begin{equation}
C_{\mathrm{front}}^{\mathrm{dyn}}(t)
=
\max_{\mathbf{q}_i\in S_p}
C_{\mathrm{dyn}}(\mathbf{q}_i,t).
\label{eq:front_dynamic_cost}
\end{equation}

The velocity and acceleration limits are adjusted according to a speed-reduction threshold $C_s$ and a stopping threshold $C_{\mathrm{stop}}$. When
$C_s\leq C_{\mathrm{front}}^{\mathrm{dyn}}<C_{\mathrm{stop}}$,
the scaling ratio is
\begin{equation}
\gamma
=
1-
\frac{C_{\mathrm{front}}^{\mathrm{dyn}}-C_s}
{C_{\mathrm{stop}}-C_s}
(1-\gamma_{\min}),
\label{eq:velocity_ratio}
\end{equation}
with $\gamma_{\min}$ the minimum scale. The limits are
\begin{equation}
v_{\max}'=\max(v_{\min},\gamma v_{\max}),
\qquad
a_{\max}'=\max(a_{\min},\gamma a_{\max}).
\label{eq:adaptive_limits}
\end{equation}

For $C_{\mathrm{front}}^{\mathrm{dyn}}<C_s$, the nominal limits are retained. When
$C_{\mathrm{front}}^{\mathrm{dyn}}\geq C_{\mathrm{stop}}$,
trajectory tracking is suspended and the UAV switches to hovering until a feasible local trajectory available.

\subsection{Guide-Driven 3D Trajectory Optimization}

At each replanning cycle, a local segment of the 2D guide is lifted to the flight altitude $h_f$:
\begin{equation}
\mathbf{q}^{\mathrm{ref}}_k
=
(x_k,y_k,h_f).
\label{eq:lifted_guide_point}
\end{equation}
The segment extends from the guide point nearest to the current planning state to a finite look-ahead target.

The local trajectory $\mathbf{p}_r(t)$ is parameterized using MINCO~\cite{wang2022geometrically}. The optimization objective is
\begin{equation}
\begin{aligned}
J={}&
J_{\mathrm{smooth}}
+\lambda_{\mathrm{obs}}J_{\mathrm{obs}}
+\lambda_{\mathrm{soft}}J_{\mathrm{soft}} \\
&+\lambda_{\mathrm{feas}}J_{\mathrm{feas}}
+\lambda_{\mathrm{var}}J_{\mathrm{var}}
+\lambda_{\mathrm{wp}}J_{\mathrm{wp}}
+\lambda_t J_t ,
\end{aligned}
\label{eq:traj_objective}
\end{equation}
where the terms account for trajectory smoothness, obstacle avoidance, dynamic feasibility, point regularization, guide following, and execution time. The guide-following term is
\begin{equation}
J_{\mathrm{wp}}
=
\sum_k
\left\|
\mathbf{p}_r(t_k)-\mathbf{q}^{\mathrm{ref}}_k
\right\|^2 ,
\label{eq:guide_waypoint_cost}
\end{equation}
which incorporates mission-level guidance into the local optimizer while allowing local deformation for obstacle avoidance and dynamic feasibility.

Before execution, the optimized trajectory is checked against the local inflated 3D occupancy map. If the minimum collision distance falls below the emergency threshold, the trajectory is rejected and replanning or hovering is triggered according to the forward dynamic risk.
\section{Experiments}

\subsection{Comparative Experiments}

Six settings are considered to evaluate the proposed framework against EGO-Swarm-v2~\cite{zhou2022swarm}, as summarized in Table~\ref{tab:comparative_settings}. The settings cover different team sizes and communication conditions. For comparison, EGO-Swarm-v2 is provided with the same aligned topological target as our method.

\begin{figure*}[t]
    \centering
    \includegraphics[width=0.95\textwidth]{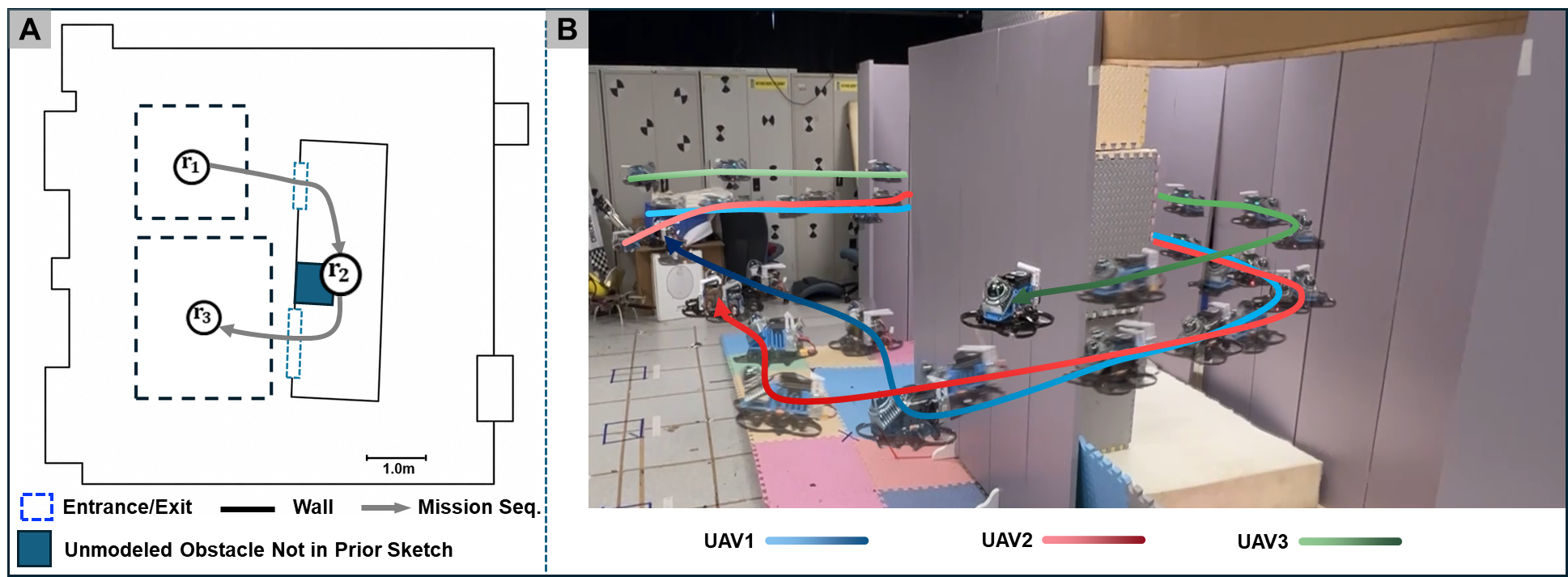}
    \caption{Real-world multi-UAV experiment in a structured indoor environment.
    (A) Experimental layout, where the dark-blue region denotes an unmodeled obstacle absent from the prior sketch.
    (B) Trajectories of three UAVs following the region-level mission sequence. Different colors represent different UAVs, while color intensity indicates temporal progression.}
    \label{fig:real_world_experiment}
\end{figure*}

The experiments are conducted in the structured multi-room environment shown in Fig.~\ref{fig:comparative_experiments}(A), consisting of six connected task regions $r_1$--$r_6$. The UAVs follow the mission sequence
$r_1 \rightarrow r_4 \rightarrow r_2 \rightarrow r_5 \rightarrow r_3 \rightarrow r_6$.
The task evaluates region-level mission execution under constrained inter-region passages and dynamic obstacle interactions.
\begin{table}[h]
\centering
\caption{Settings of the comparative experiments.}
\label{tab:comparative_settings}
\footnotesize
\setlength{\tabcolsep}{1.6pt}
\renewcommand{\arraystretch}{1.0}
\begin{tabular}{lccc}
\toprule
ID & Method & UAV Num. & Comm. \\
\midrule
P1   & Proposed     & 1 & -- \\
P2   & Proposed     & 2 & Avail. \\
P3C  & Proposed     & 3 & Avail. \\
P3N  & Proposed     & 3 & Unavail. \\
B1   & EGO-Swarm-v2 & 1 & -- \\
B3C  & EGO-Swarm-v2 & 3 & Avail. \\
\bottomrule
\end{tabular}
\end{table}

Each setting is repeated ten times. We report mission success rate (\emph{Succ.}), mission duration (\emph{Dur.}), trajectory length (\emph{Path}), minimum obstacle clearance (\emph{Clear.}), minimum inter-UAV distance (\emph{UAV Dist.}), and maximum flight height (\emph{Height}). Mission duration is measured from the start of UAV1 motion until the final UAV stops. For multi-UAV settings, \emph{Path} is first averaged across UAVs within each trial and then across repeated trials. \emph{Height} corresponds to the maximum flight height averaged over repeated trials. The fastest trajectory set for each setting is shown in Fig.~\ref{fig:comparative_experiments}(B), and the quantitative results are summarized in Table~\ref{tab:comparative_results}.

\begin{table}[h]
\centering
\caption{Quantitative results of comparative experiments.}
\label{tab:comparative_results}
\footnotesize
\setlength{\tabcolsep}{1.6pt}
\renewcommand{\arraystretch}{1.0}
\begin{tabular}{lcccccc}
\toprule
ID & Succ. & Dur. & Path & Clear. & UAV Dist. & Height \\
   &       & (s)  & (m)  & (m)    & (m)       & (m)    \\
\midrule
B1   & 9/10  & 82.60  & 72.55 & 0.288 & --    & 3.42 \\
B3C  & 5/10  & 100.61 & 73.14 & 0.222 & 0.311 & 3.40 \\
P1   & 10/10 & 59.56  & 67.78 & 0.303 & --    & 1.55 \\
P2   & 10/10 & 90.12  & 70.16 & 0.303 & 0.921 & 1.54 \\
P3C  & 10/10 & 105.17 & 72.45 & 0.303 & 0.707 & 1.54 \\
P3N  & 10/10 & 122.64 & 71.19 & 0.290 & 0.712 & 1.56 \\
\bottomrule
\end{tabular}
\end{table}

As shown in Table~\ref{tab:comparative_results}, the proposed method achieves a 10/10 mission success rate in all tested configurations. In the single-UAV comparison, P1 completes the mission in 59.56~s with an average path length of 67.78~m, compared with 82.60~s and 72.55~m for B1, which succeeds in 9/10 trials. The average maximum flight height is also reduced from 3.42~m for B1 to 1.55~m for P1. The baseline may exploit vertical motion to bypass local constraints, whereas the proposed mission-oriented guide favors motion within the task-relevant navigation layer. These results indicate reduced task-irrelevant motion and improved mission efficiency.

The difference is more pronounced in the three-UAV comparison. B3C succeeds in 5/10 trials, with a minimum inter-UAV distance of 0.311~m and obstacle clearance of 0.222~m, whereas P3C completes all ten trials with corresponding distances of 0.707~m and 0.303~m. The unsuccessful B3C trials are primarily associated with local trajectory-optimization failures in narrow and structured regions. Under such conditions, simultaneously satisfying dynamic-feasibility, obstacle-avoidance, and inter-UAV separation constraints can make local optimization sensitive to initialization. The proposed framework instead provides a mission-oriented guide derived from the aligned structural prior, offering an informative reference for local trajectory generation. The results suggest that this guidance improves both mission reliability and trajectory-generation robustness in structured multi-room environments.

The comparison between P3C and P3N further evaluates operation under communication loss. Without inter-UAV communication, P3N maintains a 10/10 success rate and a minimum inter-UAV distance of 0.712~m, while the mission duration increases from 105.17~s to 122.64~s. This increase is consistent with the reliance on local observations and more conservative dynamic-obstacle handling. The results indicate that the framework can maintain mission execution under communication-unavailable conditions, albeit with reduced execution efficiency.

\subsection{Real-World Experiments}

Real-world experiments are conducted with three physical UAVs in the environment shown in Fig.~\ref{fig:real_world_experiment}. The environment contains two open regions, $r_1$ and $r_3$, and a room-like region $r_2$ with a 0.8~m entrance and a 1.2~m exit. An obstacle absent from the prior sketch is placed inside $r_2$ to evaluate online obstacle handling. The UAVs complete the mission sequence
$r_1 \rightarrow r_2 \rightarrow r_3$
within 17~s under a 10\% communication packet-loss rate. The experiment demonstrates that the aligned sketch prior can be converted into executable multi-UAV motion while accommodating an unmodeled local obstacle and communication packet loss.

\subsection{Multi-Floor Generalization}

\begin{figure}[t]
    \centering
    \includegraphics[width=\columnwidth]{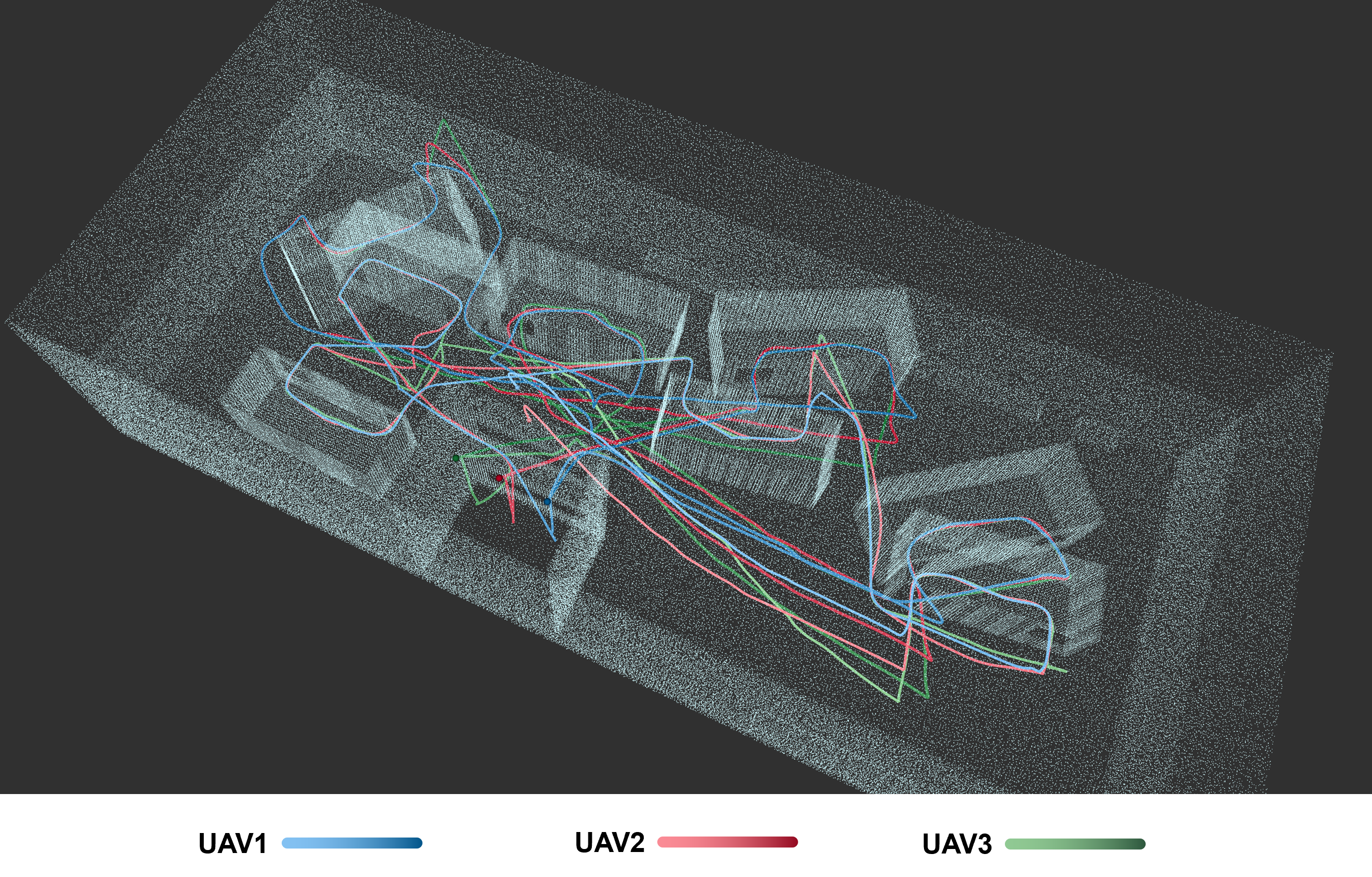}
    \caption{Multi-floor simulation in a reconstructed 3D point-cloud environment. The light-blue point cloud represents the building structure, while the blue, red, and green curves denote the trajectories of different UAVs.}
    \label{fig:multi_floor_experiment}
\end{figure}

To evaluate extension to layered building structures, a multi-floor simulation is conducted in which floor-wise planar regions are connected through designated vertical transition regions. As shown in Fig.~\ref{fig:multi_floor_experiment}, three UAVs generate continuous trajectories through the reconstructed multi-floor point-cloud environment, following region-level mission guidance across an inter-floor transition and continuing on the target floor. The result indicates that the proposed system can be extended beyond a single-floor planar layout.

\section{Conclusion}

This paper presented a mission-oriented coordinated navigation framework for multi-UAV indoor operations using sketch-map priors. The framework integrates topological alignment, static and dynamic traversability, and region-state coordination into a mission-oriented representation, which supports layered 2D-3D navigation from region-level guidance to local collision-free trajectory generation. Simulation and real-world experiments demonstrated effective mission execution across different UAV team sizes and communication conditions, including scenarios with unmodeled obstacles and communication loss. Multi-floor simulations further showed the extension of the proposed framework to layered indoor structures. Future work will investigate richer semantic mission specifications, more complex multi-floor environments, and larger-scale heterogeneous UAV teams.

\vspace{-10 pt}
\bibliographystyle{ieeetr}
\bibliography{reference}

@article{li2026fu,
  title={FU-MPC: Frontier-and Uncertainty-Aware Model Predictive Control for Efficient and Accurate UAV Exploration with Motorized LiDAR},
  author={Li, Jianping and Wan, Pengfei and Liu, Zhongyuan and Wang, Yi and Chen, Yiheng and Xu, Xinhang and Jin, Rui and Zhou, Boyu and Xie, Lihua},
  journal={arXiv preprint arXiv:2605.14920},
  year={2026}
}

@article{wang2026palm,
  title={Palm-sized Omnidirectional Vision-Based UAV Exploration with Sparse Topological Map Guidance},
  author={Wang, Zirui and Luo, Xinjia and Sun, Haotian and Ma, Jun and Guo, Jian and Zhou, Boyu},
  journal={arXiv preprint arXiv:2605.07275},
  year={2026}
}

@article{li2026aeos,
  title={Aeos: Active environment-aware optimal scanning control for uav lidar-inertial odometry in complex scenes},
  author={Li, Jianping and Xu, Xinhang and Liu, Zhongyuan and Yuan, Shenghai and Cao, Muqing and Xie, Lihua},
  journal={ISPRS Journal of Photogrammetry and Remote Sensing},
  volume={232},
  pages={476--491},
  year={2026},
  publisher={Elsevier}
}

@inproceedings{xu2024cost,
  title={A cost-effective cooperative exploration and inspection strategy for heterogeneous aerial system},
  author={Xu, Xinhang and Cao, Muqing and Yuan, Shenghai and Nguyen, Thien Hoang and Nguyen, Thien-Minh and Xie, Lihua},
  booktitle={2024 IEEE 18th International Conference on Control \& Automation (ICCA)},
  pages={673--678},
  year={2024},
  organization={IEEE}
}

@article{cao2023distributed,
  title={Distributed control of multirobot sweep coverage over a region with unknown workload distribution},
  author={Cao, Muqing and Cao, Kun and Li, Xiuxian and Xie, Lihua},
  journal={IEEE Transactions on Systems, Man, and Cybernetics: Systems},
  volume={53},
  number={10},
  pages={6503--6515},
  year={2023},
  publisher={IEEE}
}

@inproceedings{cao2020online,
  title={Online trajectory correction and tracking for facade inspection using autonomous uav},
  author={Cao, Muqing and Lyu, Yang and Yuan, Shenghai and Xie, Lihua},
  booktitle={2020 IEEE 16th International Conference on Control \& Automation (ICCA)},
  pages={1149--1154},
  year={2020},
  organization={IEEE}
}

@article{cao2025cooperative,
  title={Cooperative Aerial Robot Inspection Challenge: A Benchmark for Heterogeneous Multi-Uncrewed-Aerial-Vehicle Planning and Lessons Learned},
  author={Cao, Muqing and Nguyen, Thien-Minh and Yuan, Shenghai and Anastasiou, Andreas and Zacharia, Angelos and Papaioannou, Savvas and Kolios, Panayiotis and Panayiotou, Christos G and Polycarpou, Marios M and Xu, Xinhang and others},
  journal={IEEE Robotics \& Automation Magazine},
  year={2025},
  publisher={IEEE}
}

@article{geng2025epic,
  title={Epic: A lightweight lidar-based uav exploration framework for large-scale scenarios},
  author={Geng, Shuang and Ning, Zelin and Zhang, Fu and Zhou, Boyu},
  journal={IEEE Robotics and Automation Letters},
  year={2025},
  publisher={IEEE}
}

@article{wang2022geometrically,
  title={Geometrically constrained trajectory optimization for multicopters},
  author={Wang, Zhepei and Zhou, Xin and Xu, Chao and Gao, Fei},
  journal={IEEE Transactions on Robotics},
  volume={38},
  number={5},
  pages={3259--3278},
  year={2022},
  publisher={IEEE}
}

@article{jin2025tethered,
  title={Tethered UAV Autonomous Knotting on Environmental Structures for Transport},
  author={Jin, Rui and Xu, Xinhang and Yang, Yizhuo and Li, Jianping and Cao, Muqing and Xie, Lihua},
  journal={Cyborg and Bionic Systems},
  volume={6},
  pages={0450},
  year={2025},
  publisher={AAAS}
}

@inproceedings{jin2024gs,
  title={Gs-planner: A gaussian-splatting-based planning framework for active high-fidelity reconstruction},
  author={Jin, Rui and Gao, Yuman and Wang, Yingjian and Wu, Yuze and Lu, Haojian and Xu, Chao and Gao, Fei},
  booktitle={2024 IEEE/RSJ International Conference on Intelligent Robots and Systems (IROS)},
  pages={11202--11209},
  year={2024},
  organization={IEEE}
}

@article{tianyue2026precise,
author = {Tianyue Wu  and Guangtong Xu  and Zihan Wang  and Junxiao Lin  and Tianyang Chen  and Yuze Wu  and Zhichao Han  and Zhiyang Liu  and Fei Gao },
title = {Precise aggressive aerial maneuvers with sensorimotor policies},
journal = {Science Robotics},
volume = {11},
number = {115},
pages = {eaeb0180},
year = {2026},
doi = {10.1126/scirobotics.aeb0180},
URL = {https://www.science.org/doi/abs/10.1126/scirobotics.aeb0180},
eprint = {https://www.science.org/doi/pdf/10.1126/scirobotics.aeb0180},
}

@article{zhou2022swarm,
  title={Swarm of micro flying robots in the wild},
  author={Zhou, Xin and Wen, Xiangyong and Wang, Zhepei and Gao, Yuman and Li, Haojia and Wang, Qianhao and Yang, Tiankai and Lu, Haojian and Cao, Yanjun and Xu, Chao and others},
  journal={Science robotics},
  volume={7},
  number={66},
  pages={eabm5954},
  year={2022},
  publisher={American Association for the Advancement of Science}
}

@article{zacharia2026omniplanner,
  title={OmniPlanner: Universal Exploration and Inspection Path Planning across Robot Morphologies},
  author={Zacharia, Angelos and Dharmadhikari, Mihir and Singh, Mohit and Alexis, Kostas},
  journal={arXiv preprint arXiv:2603.04284},
  year={2026}
}

@inproceedings{tang2023bubble,
  title={Bubble explorer: Fast UAV exploration in large-scale and cluttered 3D-environments using occlusion-free spheres},
  author={Tang, Benxu and Ren, Yunfan and Zhu, Fangcheng and He, Rui and Liang, Siqi and Kong, Fanze and Zhang, Fu},
  booktitle={2023 IEEE/RSJ International Conference on Intelligent Robots and Systems (IROS)},
  pages={1118--1125},
  year={2023},
  organization={IEEE}
}

@article{dharmadhikari2025semantics,
  title={Semantics-aware predictive inspection path planning},
  author={Dharmadhikari, Mihir and Alexis, Kostas},
  journal={IEEE Transactions on Field Robotics},
  year={2025},
  publisher={IEEE}
}

@article{wu2026layer,
  title={Layer-Based Multi-Stage UAV Coverage Path Planning for Energy-Efficient Image-Based 3D Site Modeling},
  author={Wu, Zebiao and Marais, Patrick},
  journal={Unmanned Systems},
  pages={1--25},
  year={2026},
  publisher={World Scientific}
}

@article{hong2026post,
  title={Post-Disaster Multi-UAV Task Planning Based on Graph Neural Network Decoupling},
  author={Hong, Rui and Zhang, Jia and Gan, Minggang and Wang, Qing and Xin, Bin and Chen, Jie},
  journal={Unmanned Systems},
  pages={1--16},
  year={2026},
  publisher={World Scientific}
}

@inproceedings{baek2025pipe,
  title={Pipe planner: Pathwise information gain with map predictions for indoor robot exploration},
  author={Baek, Seungjae and Moon, Brady and Kim, Seungchan and Cao, Muqing and Ho, Cherie and Scherer, Sebastian and Jeon, Jeong Hwan},
  booktitle={2025 IEEE/RSJ International Conference on Intelligent Robots and Systems (IROS)},
  pages={7684--7691},
  year={2025},
  organization={IEEE}
}

@article{pan2026action,
  title={Action Correction-Enhanced Multi-Agent Reinforcement Learning for Path Planning in Urban Environments},
  author={Pan, Haixia and Han, Linfeng and Yan, Jiaming and Liu, Ruijun},
  journal={Unmanned Systems},
  volume={14},
  number={02},
  pages={461--479},
  year={2026},
  publisher={World Scientific}
}

@article{border2024osprey,
  title={Osprey: Multisession autonomous aerial mapping with LiDAR-based SLAM and next best view planning},
  author={Border, Rowan and Chebrolu, Nived and Tao, Yifu and Gammell, Jonathan D and Fallon, Maurice},
  journal={IEEE Transactions on Field Robotics},
  volume={1},
  pages={113--130},
  year={2024},
  publisher={IEEE}
}

@article{jin2026quadhand,
  title={{QuadHand}: A compact quadrotor aerial manipulator with {MRC-SDF}-based whole-body motion planning},
  author={Jin, Rui and Liu, Ruiyang and Xu, Xinhang and Jin, Haotian and Wang, Yi and Yang, Yizhuo and Xie, Lihua},
  journal={arXiv preprint arXiv:2609.35094},
  year={2026},
  url={https://arxiv.org/abs/2609.35094}
}

@article{wang2026ductam,
  title={{DuctAM}: A duct-assisted quadrotor-based aerial manipulator enabling high-force push-and-pull interactions},
  author={Wang, Yi and Jin, Rui and Xu, Xinhang and Jin, Haotian and Liu, Ruiyang and Yang, Yizhuo and Xie, Lihua},
  journal={arXiv preprint arXiv:2609.15861},
  year={2026},
  url={https://arxiv.org/abs/2609.15861}
}
\end{document}